\documentclass[sigconf]{acmart}
\AtBeginDocument{%
  \providecommand\BibTeX{{%
    \normalfont B\kern-0.5em{\scshape i\kern-0.25em b}\kern-0.8em\TeX}}}

\usepackage{xcolor}
\usepackage{tabulary}
\usepackage{adjustbox}
\usepackage{booktabs}
\usepackage{caption}
\usepackage{multirow}
\usepackage{multicol}
\usepackage{tcolorbox}
\usepackage{enumitem}
\usepackage{amsmath}
\usepackage[caption=false,font=normalsize,labelfont=sf,textfont=sf]{subfig}

\usepackage[utf8]{inputenc}
\usepackage[T1]{fontenc}
\usepackage{microtype}

\usepackage{CJK}
\usepackage{color}
\tcbset{width=0.9\textwidth,boxrule=0pt,colback=red,arc=0pt,auto outer arc,left=0pt,right=0pt,boxsep=5pt}

\newcommand{\comb}[1]{\textcolor{blue}{#1}}

\begin{document}

\title{
%
GNoM: Graph Neural Network Enhanced Language Models for Disaster Related Multilingual Text Classification
%
}
%
\author{Samujjwal Ghosh}
\email{cs16resch01001@iith.ac.in}
\orcid{0000-0003-2859-7358}
\affiliation{%
  \institution{Indian Institute of Technology Hyderabad}
  \country{India}
}

\author{Subhadeep Maji}
\affiliation{%
  \institution{Amazon}
  \country{India}
}

\author{Maunendra Sankar Desarkar}
\affiliation{%
  \institution{Indian Institute of Technology Hyderabad}
  \country{India}
}






\renewcommand{\shortauthors}{}

\begin{abstract}
  Online social media works as a source of various valuable and actionable information during disasters. These information might be available in multiple languages due to the nature of user generated content. An effective system to automatically identify and categorize these actionable information should be capable to handle multiple languages and under limited supervision. However, existing works mostly focus on English language only with the assumption that sufficient labeled data is available. To overcome these challenges, we propose a multilingual disaster related text classification system which is capable to work undervmonolingual, cross-lingual and multilingual lingual scenarios and under limited supervision. Our end-to-end trainable framework combines the versatility of graph neural networks, by applying over the corpus, with the power of transformer based large language models, over examples, with the help of cross-attention between the two. We evaluate our framework over total nine English, Non-English and monolingual datasets invmonolingual, cross-lingual and multilingual lingual classification scenarios. Our framework outperforms state-of-the-art models in disaster domain and multilingual BERT baseline in terms of Weighted F$_1$ score. We also show the generalizability of the proposed model under limited supervision.
\end{abstract}




\maketitle

\section{Introduction}
People affected by disasters turn to online social network to seek help and report actionable information. Identification and categorization of these actionable information can help in planning rescue and relief operations effectively. However, these user-generated contents, such as tweets, are generally in languages native to the location of the disaster. On the other hand, majority of works in the literature focus mostly on English language~\cite{glen, qcri, ghosh2020semi,  Mazloom2018ClassificationOT, Neppalli2018DeepNN, Nguyen2017RobustCO, caragea2016identifying} only. Understanding and processing texts in multiple languages is of paramount importance for effective disaster mitigation. A multilingual system capable of working with various languages will expand the applicability of such systems towards rescue and relief operations. Because of this, there is a strong need for a multilingual text classification framework which can identify and categorize useful and actionable information generated during disasters. Another challenge in building such an automated system is the lack of sufficient labeled data~\cite{glen, qcri} in disaster mitigation domain. Labeling examples during an ongoing disaster is costly and might not be feasible. This bottleneck becomes even more prevalent in multilingual scenario. 

Keeping the above-mentioned constraints in mind, we propose a
\textbf{G}raph \textbf{N}eural netw\textbf{o}rk based \textbf{M}ultilingual text classification framework (\textbf{GNoM})
which can work efficiently under limited labeled data in vmonolingual, cross-lingual and multilingual lingual settings. Our proposed approach enhances the power of transformer based large language models with the help of a Graph Neural Network (GNN) based formulation which enables the model to work both in multilingual setting and under limited labeled data by utilizing a word graph constructed from available textual corpus. GNoM has three main components, a Text Representer (TR), a Graph Featurizer (GF) and an Importance Estimator (IE), all of which are trained jointly in end-to-end manner.
%
The purpose of the TR is to represent mono and multi lingual texts effectively which captures example level context for better class separability. On the other hand, the GF captures corpus level context from multilingual data which enables the framework to work in both mono and cross lingual settings. Both of these components are agnostic to any specific architecture and can be realized using any transformer and GNN architectures for TR and GF respectively. This flexibility allows for easier incorporation of new and powerful architectures in future. The IE combines the two components by estimating cross attention between them.

Due to recent success of GNNs in multiple domains~\cite{yao2019graph, Wu_Lian_Xu_Wu_Chen_2020}, we explore GNN-based GF to encode relationships among words in the dataset. We construct a word graph by connecting words present in the whole corpus (i.e. labeled and unlabeled data available from dataset(s)). This word graph helps in two aspects, a) connects related and relevant words from multiple languages and b) extends the framework's capability to capture context from easily available unlabeled data. The GF component enables the framework to work with multilingual data under limited supervision by capturing prior information from the neighborhood of words.
%
The graph is carefully constructed keeping multilinguality in consideration. The GNN applied on this word graph learns a joint embedding across the languages. These word representations are passed through an importance estimator component to boost/attenuate the representations accordingly. These representations are concatenated with the representations obtained from the TR to obtain the final representation of the word. This concatenated representation projects the words to a new embedding space where words with similar context from different languages are projected closer to each other. The classifier module takes these representations as input to predict the class.


Our proposed framework outperforms state-of-the-art (SOTA) methods in disaster domain in mono, cross and multi lingual experiments.
In summary, our contributions are as follows:
\begin{itemize}
    \item We propose a framework for disaster related text classification which works across monolingual, cross-lingual and multilingual settings.
    \item The proposed framework is effective in utilizing easily available unlabeled data. At the same time, flexible with the architectures that can be used.
    \item We show significant improvement in total $9$ English Non-English and multilingual disaster related tweet classification dataset. Additionally we show that our framework is able to generalize under limited supervision.
\end{itemize}
\section{Related Works} \label{related}
There are many studies focusing on disaster-related tweet classification in both binary~\cite{Mazloom2018ClassificationOT, Neppalli2018DeepNN, Nguyen2017RobustCO, caragea2016identifying} and multi-label~\cite{glen, iaid} settings. However, much of the literature is focused on monolingual corpora, particularly, English language only~\cite{Mazloom2018ClassificationOT, Neppalli2018DeepNN, Nguyen2017RobustCO, caragea2016identifying}. Whereas, in a real-world scenario, user generated information may come in any language. We first explore the literature of multilingual learning in disaster response followed by approaches which focus on incorporating an additional graph component.
    
There are some notable studies which look into the multilingual direction. We highlight a few works which explore disaster-related tweet classification in multilingual setting.
One of the comprehensive works in this area was done by Raychowdhury et al. in~\cite{raychowdhury2020cross} which explore disaster-related text classification by applying Manifold Mixup~\cite{Verma19manifoldmixup} on mBERT. They aggregated multiple disaster datasets containing tweets in multiple languages into a single large dataset and performed their experiments on that dataset. 
Krishnan et al.~\cite{Krishnan2020AttentionRA} explored classification of crisis related tweets using attention realignment by introducing a language classifier in addition to task classifier. They use XLM-Roberta architecture as the multilingual featurizer. However, their approach is dependent on availability of parallel corpora. Piscitelli et al. explored application of context-independent multilingual word embeddings called MUSE~\cite{conneau2017word} to perform tweet classification during emergencies in their work~\cite{Piscitelli2021multilingual}. Similarly, Lorini et al. used context-independent multilingual embeddings for their flood recognition system based on online social media called European Flood Awareness System (EFAS) in~\cite{lorini2019integrating}. However, context-independent word embedding typically fails to capture relevant information~\cite{Barua21analysis}.
Torres et al. explored crisis-related conversations in a cross-lingual setting~\cite{torres2019cross}. Their study was limited to Spanish and English tweets only. The work~\cite{Musaev2017TowardsMA} by Musaev et al., filters tweets relevant to landslides using Wikipedia articles as knowledge repository. One limitation of this approach is that the model needs the same Wikipedia article in multiple languages to learn multilingual embeddings. In~\cite{khare2018cross}, Khare et al. classify relevancy of tweets from 30 crisis events in 3 languages (i.e. English, Spanish, and Italian). The main drawback in their approach is that it is limited to the languages present in the training data only and does not generalize to other languages.

There are a few approaches which tried to incorporate a graph component to the model in the domain of disaster management. These approaches majorly differ in how the graph is formed and what kind of information they are trying to capture. However, none of the approaches explored the multilingual perspective.
Alam et al. proposed end-to-end approach based on adversarial learning in their work~\cite{qcri} (DAAT). They employ a GNN based component to construct a document-level graph by calculating k-nearest neighbours using Word2Vec~\cite{word2vec} vectors. 
Li et al. use a Domain Reconstruction Classification
Network (DRCN) in their work~\cite{sigir'20} for disaster related text classification. DRCN reconstructs the target domain data with an autoencoder to minimize domain shift. Both DAAT and DRCN approaches are designed for domain adaptation setting in English language only.
Zahera et al. explored the combination of GAT~\cite{gat} with mBERT~\cite{bert} encoder in~\cite{iaid}. However, their graph formulation is heterogeneous and based on the implicit assumption that sufficient training data is available which might not be true in disaster scenarios. They incorporate class labels in addition to named entities as additional nodes during graph construction. Additionally, their work is focused only on English language.
Ghosh et al. proposes a 2-part global and local graph neural network based technique called GLEN~\cite{glen} to utilize global token graph to learn domain agnostic features. They used token as nodes in the graph with token cooccurrence as edges. This graph construction is somewhat similar to ours. However, their formulation is limited to monolingual (English) setting only. We compare GNoM with GLEN in monolingual experiments (Ref. Section~\ref{sec:experiments}).
\begin{figure*}
\includegraphics[width=\textwidth]{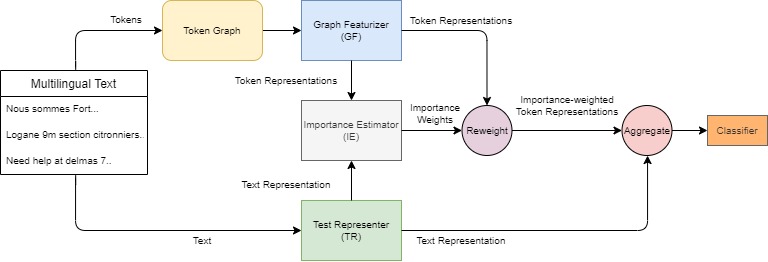}
\caption{Overview of our proposed GNoM framework. The ``Text Representer'' (TR)~(\S~\ref{subsection:transformer_example_context}) takes multingual examples as input and generates both example level and word level representation. It primarily captures example level context which aids in effective separation of classes.
On the other hand, tokenized multilingual unique words are used to construct the word graph which passes through the ``Graph Featurizer'' (GF)~(\S~\ref{subsection:graph_corpus_context}) and serves as a prior over the words. The ``Important Estimator''~(\S~\ref{subsection:cross_mha}) 
estimates the importance of word priors with respect to the input example by taking example representation from TR and node (word) vectors from GF.
Finally, these attention weighted node vectors are aggregated with word vectors generated by the TR and passed to the classifier.} \label{fig:diagram}
\end{figure*}


\section{Proposed Framework} \label{framework}
Our method comprises of three main components which attempt to capture the corpus level (e.g dataset) and example level (e.g tweet) context information representing same word in two different embedding spaces and an attention mechanism to decide the importance of those embeddings. The corpus-level information serves as prior for word representation by aggregating over many contexts in which a word appears while the example-level information captures the context specific to an example for better class separation. We claim that modelling the two contexts explicitly helps in better generalization in a low resource text classification setting such as disaster response, as compared to using example level context (by employing, for example, transformer based models) alone. We establish this claim empirically in our experiments (\S~\ref{sec:experiments}). We combine the two embedding spaces with a novel end-to-end scaled dot product cross attention mechanism which learns to attend on corpus level context information given an example, where the downstream task is text classification. In addition, we enable multilinguality in our model by making it applicable to realistic disaster situations while most existing works on disaster response domain are monolingual~\cite{glen, qcri, ghosh2020semi} only. 

In \S~\ref{subsection:transformer_example_context}, we discuss our method for obtaining example level contextual word embeddings with recent transformer based models (e.g BERT~\cite{bert} or XLM-R~\cite{conneau2019unsupervised}). Our method is naturally multilingual by virtue of using a multilingual transformer model as the Text Representer. 
In \S~\ref{subsection:graph_corpus_context}, we discuss our method for obtaining corpus level word embeddings with a Graph Featurizer (GF). We base the featurizer on Graph Convolution Network (GCN) on a word graph constructed from available labeled and unlabeled data. Our word graph is multilingual containing nodes (tokens) from multiple languages, and defines edges using embedding similarity between cross-language words from pretrained multilingual transformer models. In \S~\ref{subsection:cross_mha}, we discuss our method for combining the corpus level and example level word embeddings using a scaled dot product attention scheme, called Importance Estimator, which uses similarity between example embedding and individual GCN node embeddings to the compute attention scores. 

We will refer to examples (e.g. tweets) as $\mathbf{x}=(x_1, x_2,\ldots x_n)$ where $x_i$ is the $i^{\mathtt{th}}$ word. We assume access to labeled dataset for every classification task; $\mathcal{L} = \{(\mathbf{x}, y)\}$, where $y$ is binary, multi-class or multi-label target depending on the task. In addition, for some tasks, we also utilize unlabeled data $\mathcal{U} = \{(\mathbf{x})\}$ when available (details on the collection process is deferred to \S~\ref{sec:experiments}) which we use in the construction of the word graph while learning corpus level context embeddings (\S~\ref{subsection:graph_corpus_context}). Focus of the current work is multilingual text classification in disaster domain which is typically low resource, therefore our labeled datasets are small (on average $\approx5\mathtt{K}$ labeled examples).

\subsection{Example Level Context Embedding} \label{subsection:transformer_example_context}
We employ a transformer based model as the Text Representer to learn example level contextual word embeddings. Transformer models owing to their self-attention structure learn word embeddings for a word depending on its \textit{similarity} to all the other words in the example. The main objective of the text representer is to represent multilingual text effectively, and at the same time learn an embedding space which increases the separation among the classes. In the monolingual setting we use BERT and in multilingual setting we use mBERT to represent examples. In both settings, the pooled token embedding (i.e. [CLS] for BERT) is considered as the example embedding. The [CLS] token embedding based text representation has been widely used for downstream classification tasks~\cite{wang2021transformer}. We would like to emphasize that our overall model is not tied to BERT architecture and can be replaced with any transformer based text representation model architecture, for example XLM-RoBERTa. In context of an example $\mathbf{x}$, we will denote the embedding of $\mathbf{x}$ as $[\mathtt{CLS}]_{\mathbf{x}}$ and individual words $x_i$ as $\mathbf{h}_{\mathbf{T}\mid \mathbf{x}}(x_i)$. 

%
%


\subsection{Corpus Level Context Embedding} \label{subsection:graph_corpus_context}

We propose a graph neural network based Graph Featurizer to learn corpus level context based word embeddings. We define a word graph whose vertices are unique words $x_i$ from examples $\mathbf{x} \in \mathcal{L} \cup \mathcal{U}$. In some tasks there is no unlabeled dataset (i.e $\mathcal{U}=\phi$). Typically edges in word graphs are defined purely in terms of co-occurrence (within a window) of words from examples of an underlying corpus~\cite{glen}. However, this fails to capture multilinguality because words from different languages seldom co-occur in a example, which in turn will result in a word graph with disconnected components. 
To address this limitation, we obtain embedding similarity from embedding layer of the transformer based large multilingual language model. 
We will refer to co-occurrence (within a window in examples) based similarity as matrix $C_{i,j}$ and embedding based similarity as matrix $E_{i,j}$. Matrices $C$ and $E$ are row-normalized and added (i.e $S=C+E$) to obtain the combined measure of similarity. The similarity values above a threshold ($S_{i,j} > \tau$) are used to define edges in the graph. The threshold ($\tau$) is a hyperparameter in our model. As a pre-processing step, very infrequent words (minimum corpus frequency of $3$) and high frequency stopwords are not considered as nodes in the word graph. Co-occurrence similarity helps expand context over words within a language, whereas embedding similarity captures relationship across words from multiple languages.
We initialize the word graph's initial embeddings with the word embedding layer word representation from the Text Representer. This initialization technique serves two advantages: (a) as the multilingual TR's are generally pretrained with large corpus we are able to bring this prior information in the formulation of the word graph and (b) enables both the Text Representer and the Graph Featurizer to have the same vocabulary. On the word graph, we apply a $k$-hop GCN to obtain graph based token (node) embeddings. GF expands the context information present in immediate neighborhood of nodes (i.e frequently co-occurring words/high embedding similarity) in the graph to its $k$-hop neighborhoods by aggregating information over multiple hops. A high value of $k$ expands the context to a larger neighborhood but risks oversmoothing~\cite{oono2020graph}, whereas smaller $k$ will limit the context expansion. We set $k=2$ in all our experiments. We will denote the graph based embedding of word (node) $v$ as $\mathbf{h}_{\mathbf{G}}(v)$.


\subsection{Scaled Dot Product Cross Attention} \label{subsection:cross_mha}
We now turn to the question of combining the above two embedding spaces to improve generalization of the overall model on downstream text classification task. The graph based embedding of a word $x_i$ ($\mathbf{h}_{\mathbf{G}}(x_i)$) in an example $\mathbf{x}$ is independent of rest of words in $\mathbf{x}$ and is based on information propagation over its $k$-hop graph neighborhood. Therefore, the graph based embedding serves as a \textit{prior} for word representation. We propose to combine this prior information with example level embedding $\mathbf{h}_{\mathbf{T}\mid \mathbf{x}}(x_i)$ using a scaled dot product attention which chooses to (ignore)attend to a prior word representation basis how (dis)similar the prior is to the pooled example embedding $[\mathtt{CLS}]_{\mathbf{x}}$. These attention scores works as an Importance Estimator in context to the example. In the standard scaled dot-product attention notation, the query ($Q$) is $[\mathtt{CLS}]_{\mathbf{x}}$ and keys ($K$) and values ($V$) are both node vectors corresponding to words $x_{i=1\ldots n}$ in $\mathbf{x}$. Formally, 
    \begin{align*}
        Q & = [\mathtt{CLS}]_{\mathbf{x}} \\ 
        K & = V = (\mathbf{h}_{\mathbf{G}}(x_i))_{i=1\ldots n} \\
        A(Q,K,V;W) & = \mathtt{Softmax} \left (\frac{(W_qQ) (W_kK^{T})}{\sqrt{d}} \right) 
    \end{align*}
    Here $W_q,W_k$ are parameters of the dot-product attention. The attention scores $A_{i=1\ldots n}$ form a distribution over values $V$. We combine the two embeddings by concatenating the attention multiplied prior embedding with example level context embedding per word as follows $[A_i*\mathbf{h}_{\mathbf{G}}(x_i);\mathbf{h}_{\mathbf{T}\mid \mathbf{x}}(x_i)]$. We refer to this mechanism as scaled dot product cross attention because embeddings from one subspace (example level context) serve as query for computing attention on another subspace (corpus level context). The attention layer is learned end-to-end with a classification task. In our experiments, we show an ablation study against the naive strategy of simply concatenating the two embeddings and establish the effectiveness of our scheme.    
%
\section{Experimental Setup} \label{sec:experiments}
Our goal is to build a disaster-related text classification system which works acrossvmonolingual, cross-lingual and multilingual lingual settings. 
Particularly, we aim to answer the following research questions via our experiments:
\begin{itemize}
    \item How does the performance of GNoM compare to state-of-the-art mono/cross/multi lingual models in disaster-related text (e.g., tweets) classification domain?
    \item Is GNoM capable of working when the amount of training data available is very limited?
    \item How does each component of GNoM impacts classification performance (i.e., Ablation Study)?
    %
\end{itemize}
%
In disaster domain, it is imperative that the system works under limited supervision. 
To verify the effectiveness of GNoM in such scenarios, similar to~\cite{glen}, we reduce the training data to $50\%$, $25\%$ and $10\%$ of the original training set without changing the validation and test sets.
\subsection{Datasets} \label{subsec:datasets}
We performed experiments on total $9$ datasets, out of which $5$ are in English, $3$ are in Non-English (e.g. Spanish, Italian, etc.) language and $1$ contains multilingual data. All the datasets are publicly available containing disaster related tweets. 
To perform experiments in both in-domain and cross-domain settings, we pair up datasets with same class labels.
\subsubsection{English Datasets} \label{subsec:english_datasets}
For experiments with English language, we used publicly available two binary datasets and two multi-label datasets of tweets generated during disasters. The binary datasets `2013 Queensland Flood' (QFL) and `2015 Nepal Earthquake' (NEQ)~\cite{qcri} are labeled with relevance of tweets as classes. We present the class specific details of these two datasets in Table~\ref{table:binary_details}. The unlabeled part of both the datasets were downloaded using Twitter's public API. We obtained a total of $49,223$ and $15,464$ tweets from NEQ and QFL datasets respectively. 
We used the train, dev and test split provided by the authors as train, validation and test set.
\begin{table}[ht]
\centering
    \begin{tabular}{@{}llrr|rrr@{}}
    \toprule
     \textbf{Dataset} & \textbf{Language} & \textbf{1} & \textbf{0} & \textbf{Train} & \textbf{Val} & \textbf{Test} \\ \midrule
    QFL & English & 5414 & 4619 & 6019 & 1003 & 3011 \\ 
    NEQ & English & 5527 & 6141 & 7000 & 1166 & 3502 \\
    ChileEQT1 & Spanish & 928 & 1259 & 1312 & 88 & 787 \\
    SoSItalyT4 & Italian & 4739 & 903 & 3385 & 226 & 2031 \\
    EcuadorS & Spanish & 2322 & 1846 & 2501 & 167 & 1500 \\
    EcuadorE & English & 2249 & 1946 & 2515 & 180 & 1500 \\
    \bottomrule
    \end{tabular}
\caption{Details of QFL, NEQ, ChileEQT1, SoSItalyT4 and Ecuador datasets. 1 and 0 indicate relevant and irrelevant classes.}
\label{table:binary_details}
\end{table}

\begin{table}[htbp]
\centering
\begin{adjustbox}{width=\linewidth}
\begin{tabular}{@{}lccl@{}}
\toprule
\multicolumn{2}{c|}{FIRE16} & \multicolumn{2}{c}{SMERP17} \\ \midrule
\textbf{Title} & \textbf{Class} & \textbf{Class} & \textbf{Title} \\ \midrule
Resources Available          & 1           & \multirow{2}{*}{1} & \multirow{2}{*}{Resources Available} \\ 
Medical Resources Available  & 3           &                    &                                      \\ \midrule
Resources Required           & 2           & \multirow{2}{*}{2} & \multirow{2}{*}{Resources Required}  \\ 
Medical Resources Required   & 4           &                    &                                      \\ \midrule
Resources Specific Locations & 5           & -                  &                                      \\ \midrule
Infrastructure Damage \& Restoration & 7 & 3 & Infrastructure Damage \& Restoration                    \\ \midrule
Activities NGOs / Government & 6           & 4                  & Rescue Activities NGOs / Government \\ \bottomrule
\end{tabular}
\end{adjustbox}
\caption{ Class mapping from FIRE16 to SMERP17. Class 5 of FIRE16 was ignored.}
\label{table:smerp17_fire16_map}
\end{table}

We also experiment with two multi-label datasets, namely, `Forum for Information Retrieval Evaluation 2016' (FIRE16)~\cite{fire16} and `Social Media for Emergency Relief and Preparedness' (SMERP17)~\cite{smerp17}, containing tweets collected during Nepal 2015 earthquake and 2016 Italy earthquake respectively. Tweets in these datasets are labeled with multi-label annotation where each example may belong to one or more classes. FIRE16 dataset has seven classes and SMERP17 contains four classes. We pair up FIRE16 and SMERP17 datasets according to the mapping used in~\cite{glen}. We used the mapped FIRE16 dataset for all our experiments related to FIRE16 dataset. We also collected $68,964$ unlabeled tweets for the SMERP17 dataset. Note that NEQ and FIRE16 dataset refers to the same 2015 Nepal earthquake with binary and multi-label annotations respectively. We used the same unlabeled set for both datasets. Details of the FIRE16 and SMERP17 datasets are provided in Table~\ref{table:multi_datasets}.
%
%
\begin{table*}
\begin{tabular}{@{}lrrr@{}}
    \toprule
    \textbf{Class} & \textbf{Train} & \textbf{Val} & \textbf{Test} \\ \midrule
    1 & 498 & 55  & 237 \\
    2 & 217 & 24  & 104 \\
    3 & 367 & 41  & 175 \\
    4 & 302 & 34  & 144 \\ \midrule
    Example Count & 957 & 106 & 459 \\ \bottomrule
\end{tabular}
\hfill
\begin{tabular}{@{}lrrr@{}}
    \toprule
    \textbf{Class} & \textbf{Train} & \textbf{Val} & \textbf{Test} \\ \midrule
    1 & 184 & 22  & 76 \\
    2 & 105 & 15  & 46 \\
    3 & 774 & 141 & 393\\
    4 & 212 & 25  & 80 \\ \midrule
    Example Count & 1159 & 189 & 548 \\ \bottomrule
\end{tabular}
\hfill
\begin{tabular}{@{}lrrr@{}}
    \toprule
    \textbf{Class} & \textbf{Train} & \textbf{Val} & \textbf{Test} \\ \midrule
    1  & 5933 & 544 & 861 \\
    2  & 3409 & 304 & 509 \\
    3  & 1328 & 144 & 240 \\
    4  & 18132 & 1635 & 2197 \\
    5  & 17865 & 1599 & 2121 \\
    \midrule
    Total & 46667 & 4226 & 5928 \\ \bottomrule
\end{tabular}
\caption{Details of FIRE16 (left), SMERP17 (middle) and MixUp (right) datasets. FIRE16 and SMERP17 contains tweets in English with multi-label annotation whereas MixUp contain tweets in multiple languages with only multi-class annotation. We provide example count for multi-label datasets as it may differ from the total number of annotations.}
\label{table:multi_datasets}
\end{table*}





%
%
\subsubsection{Non-English Datasets} \label{subsec:nonenglish_datasets}
We use four datasets collected from different sources. 
Ray Chowdhury et al. curated a large multilingual dataset of $134420$ tweets~\cite{raychowdhury2020cross}, annotated with five classes in multi-class setting, related to multiple disasters. They provided the train, validation and test split of the data in the form of tweet ids, due to Twitter's policy on data sharing, which we tried to download. However, we could only download $46667$, $4226$ and $5928$ tweets from the train, validation and test sets respectively. Table~\ref{table:multi_datasets} contains details about the dataset, named `MixUp'.
The dataset `Ecuador' was collected by Torres et al. in~\cite{torres2019cross}. The dataset is a collection of two datasets containing $4195$ tweets in English and $4168$ tweets in Spanish language, generated during Ecuadorian Earthquake in April 2016. We refer these two collections as EcuadorE for English and EcuadorS for Spanish collection. Both the collections contain binary annotation. We pair up the datasets for our cross lingual experiments. Details about the dataset is available in Table~\ref{table:binary_details}.
We also collected two datasets from the CrisisLex platform namely ChileEarthquakeT1~\cite{cobo15chile}, and SOSItalyT4~\cite{cresci15italy}. ChileEarthquakeT1 dataset, denoted as ChileEQT1 in our experiments, is a dataset with tweets in Spanish language from the Chilean earthquake of 2010  where all the tweets are annotated with relevance. The SOSItalyT4 dataset contains tweets from four different natural disasters in Italy between 2009 and 2014. The tweets in the dataset are annotated with ``damage'', ``no damage'', or ``not relevant''. However, similar to~\cite{khare2018cross}, we convert the annotations to binary relevance with ``damage'' and ``no damage'' both indicating relevance. We pair up ChileEQT1 and SOSItalyT4 for our cross lingual experiments.
%
%
%
\subsection{Baselines} \label{subsec:baselines}
%
Our framework enhances the transformer based Text Representer by incorporating the Graph Featurizer and Importance Estimator. To verify the effectiveness of these components, we define the vanilla Text Representer as the baseline for our experiments. 
We also compare with SOTA methods from disaster related text classification domain. A GNN based SOTA method was applied on QFL, NEQ, FIRE16 and SMERP17 in paper~\cite{glen} (GLEN) by Ghosh et al., we compare with this method in our experiments over those datasets. A few other SOTA methods presented in~\cite{qcri} (DAAT) by Alam et al. and~\cite{sigir'20} (DRCN) by Li et al. also experimented with QFL and NEQ datasets, we compare against them. 
Torres et al. in~\cite{torres2019cross} applied their approach (CLP) in both mono and cross lingual setting for Ecuador dataset. We compare with CLP in addition to vanilla mBERT for experiments over Ecuador dataset.

Recall that GNoM is flexible with the transformer architecture in the TR component. We experiment with three realizations of TR using BERT (GNoMB) for English datasets, and using mBERT (GNoMM) and XLM-RoBERTa (GNoMX) architecture for Non-English or multilingual datasets. BERT-base-uncased, BERT-Base-Multilingual-Cased and XLM-RoBERTa-Base variant are used for experiments with GNoMB, GNoMM and GNoMX respectively.
We initialize the word graph node vectors with the word embeddings of the corresponding TR. 

For our ablation study, we report results on the following ablations:
\begin{itemize} 
    \setlength\itemsep{0em}
    \item \textbf{Only TR (Without GF and IE)}: This setting corresponds to training the TR only i.e. BERT for English and mBERT for other language datasets. Only word vectors are passed to the classifier without the node vectors in the Figure~\ref{fig:diagram}.
    \item \textbf{TR+GF (Without IE)}: In this ablation, we estimate the need of Importance Estimator in our framework. 
    Both TR and GF are trained but without the IE, i.e. vectors from TR and GF are simply concatenated directly without reweighting GF vectors.
    \item \textbf{TR+GF-e+IE (Without embedding similarity edges)}: We construct the edges in the word graph using only cooccurrence for monolingual and both cooccurrence and embedding similarity for cross and multilingual settings. However, this ablation verifies the situation when only cooccurrence edges are used in cross and multi lingual settings.
    \item \textbf{GNoM Framework (GNoM)}: This setting represents our framework GNoM. We argue that GNoM is flexible with various tranformer architectures. We show two realisations of TR using (m)BERT~\cite{bert} and XLM-RoBERTa~\cite{conneau2019unsupervised} architectures for Non-English experiments.
\end{itemize}
\subsection{Training Configuration} \label{subsec:training_config}
A $2$-layer bi-directional LSTM (BiLSTM) network with a fully connected layer head is used as the classifier. Our framework was trained jointly with the classifier in an end-to-end manner. We update all the layers during training for both GNoM and the baselines. We ran each experiment $3$ times and report the average of those runs. Weighted F$_1$ score is used as the evaluation metric as it is a commonly used metric in the literature.

A few datasets have unlabeled data available in addition to labeled data. GNoM is capable to incorporate such extra data during the construction of the word graph. We utilized the unlabeled data available with QFL, NEQ, FIRE16 and SMERP17 datasets to construct the word graph. For SoSItalyT4, ChileEQT1 and Ecuador datasets, we treat the target domain train data as the unlabeled data during cross domain experiments. Note that target domain class information is not used in any of our experiments. For in-domain monolingual experiments for SoSItalyT4, ChileEQT1, Ecuador and MixUp dataset, no unlabeled data was used. For monolingual experiments, we construct the word graph using only cooccurrence, similar to~\cite{glen}, as there is no need to model inter-language relations. However, for cross and monolingual experiments we use both cooccurrence and embedding similarity to construct the edges.

We tuned our hyperparameters such as embedding similarity threshold (Ref.~\ref{subsection:graph_corpus_context}, $\tau$), learning rate and the number of epochs using the validation data. 
We searched the value of embedding similarity threshold based on performance on validation data and set the value to $0.5$ across all experiments.
We searched learning rate values with $10^{-i}$ where $i \in \{4,5,6\}$; $i=5$ found to be most suitable in majority of the training scenarios. 
%
%
\section{Results} \label{sec:results}
GNoM utilizes corpus as well as example level context to capture relations across languages. We validate the effectiveness of GNoM through multiple experiments in mono, cross and multi lingual settings.


\subsection{Monolingual Classification} \label{subsec:mono_classification}

In this setting, we use data from a single language for both training and evaluation. 
We present our findings in Tables~\ref{table:neq_qfl_f},~\ref{table:fire16_smerp17_f},~\ref{table:t1_t4_f} and ~\ref{table:ecuador_f} for QFL, NEQ, FIRE16, SMERP17, SoSItalyT4, ChileEQT1 and Ecuador datasets respectively.
For QFL and NEQ datasets (Table~\ref{table:neq_qfl_f}), we compare with GLEN, DRCN and DAAT from disaster related text classification literature and with BERT baseline. We perform experiments in both in and cross domain monolingual setting. GNoM is able to outperform GLEN (best performing among SOTA) by average $4\%$ in F$_1$ score.
We compare with GLEN and BERT for multi-label monolingual datasets FIRE16 and SMERP17 in Table~\ref{table:fire16_smerp17_f}. Our framework boosts F$_1$ significantly by as much as $6.42\%$ on average.

In Non-English SoSItalyT4, ChileEQT1 and Ecuador datasets, bottom two rows signify monolingual setting in Tables~\ref{table:t1_t4_f} and~\ref{table:ecuador_f}. No unlabeled extra data was used for these experiments. We compare with BERT baseline for SoSItalyT4, ChileEQT1 datasets. In addition, we compare with CLP for Ecuador dataset. Our approach is able to outperform BERT baseline in all $4$ scenarios. Although the performance improvement is marginal over GLEN but we achieve a significant improvement over BERT. 
%
%
%
\begin{table}[htbp]
\centering
\begin{tabular}{@{}llrrrrr@{}}
\toprule
\textbf{Source} & \textbf{Target} & \textbf{DAAT} & \textbf{DRCN} & \textbf{BERT} & \textbf{GLEN} & \textbf{GNoMB} \\ \midrule
NEQ & QFL & 65.90 & 81.18 & 80.72 & \underline{83.42} & \textbf{86.68} \\
QFL & NEQ & 59.50 & 68.38 & 67.22 & \underline{71.61} & \textbf{71.73} \\ 
\cmidrule(r){1-2}
NEQ & NEQ & 65.11 & - & 76.39 & \underline{77.76} & \textbf{78.95} \\
QFL & QFL & 93.54 & - & 96.24 & \textbf{96.77} & \underline{96.26} \\ \bottomrule
\end{tabular}
\caption{Weighted F$_{1}$ scores over NEQ and QFL datasets. 
GNoM outperforms other SOTA methods in both cross and in domain setting.}
\label{table:neq_qfl_f}
\end{table}
\begin{table}[htbp]
\centering
\begin{tabular}{@{}llrrrr@{}}
\toprule
\textbf{Source}  & \textbf{Target} & \textbf{BERT} & \textbf{GLEN} & \textbf{GNoMB} \\ \midrule
FIRE16  & SMERP17 & 76.21 & \textbf{80.80} & \underline{80.49} \\
SMERP17 & FIRE16  & 55.52 & \underline{56.56} & \textbf{62.57} \\ 
\cmidrule(r){1-2}
FIRE16  & FIRE16  & 77.36 & \underline{82.04} & \textbf{84.39} \\
SMERP17 & SMERP17 & 91.68 & \underline{93.37} & \textbf{98.15} \\ \bottomrule
\end{tabular}
\caption{Scores (Weighted F$_{1}$) of FIRE16 and SMERP17 datasets.}
\label{table:fire16_smerp17_f}
\end{table}
%
%
%
\begin{table}[htbp]
\centering
\begin{tabular}{@{}llrrrr@{}}
\toprule
\textbf{Source}  & \textbf{Target} & \textbf{mBERT} & \textbf{GNoMX} & \textbf{GNoMM} \\ \midrule
ChileEQT1 & SoSItalyT4  & 43.17 & \textbf{51.81} & \underline{49.14} \\
SoSItalyT4 & ChileEQT1  & 54.46 & \textbf{66.47} & \underline{63.20} \\
\cmidrule(r){1-2}
ChileEQT1  & ChileEQT1  & 85.32 & \underline{86.17} & \textbf{86.58} \\
SoSItalyT4 & SoSItalyT4 & 85.50 & \underline{85.64} & \textbf{85.73} \\ \bottomrule
\end{tabular}
\caption{ Weighted F$_{1}$ scores for ChileEQT1 (Spanish) and SoSItalyT4 (Italian) datasets.}
\label{table:t1_t4_f}
\end{table}
\begin{table}[htbp]
\centering
\begin{tabular}{@{}llrrrrrrr@{}}
\toprule
\textbf{Source}  & \textbf{Target} & \textbf{mBERT} & \textbf{CLP} & \textbf{GNoMX} & \textbf{GNoMM} \\ \midrule
EcuadorE & EcuadorS & 77.93 & 77.49 & \textbf{81.89} & \underline{81.54} \\
EcuadorS & EcuadorE & 90.45 & 85.88 & \textbf{91.72} & \underline{91.45} \\ 
\cmidrule(r){1-2}
EcuadorE & EcuadorE & 94.23 & 94.05 & \underline{94.30} & \textbf{94.50} \\
EcuadorS & EcuadorS & 85.18 & 85.77 & \underline{86.79} & \textbf{86.86} \\ \bottomrule
\end{tabular}
\caption{ Weighted F$_{1}$ scores between EcuadorE (English) and EcuadorS (Spanish) collections of Ecuador dataset.}
\label{table:ecuador_f}
\end{table}
\subsection{Crosslingual Classification} \label{subsec:cross_classification}
In crosslingual setting, we evaluate the classifier with data from languages that were not used during training. For example, we evaluate on Spanish data when the classifier was trained with English language. Ecuador dataset collection falls into such category when we have two collections of tweets in English and Spanish from the same disaster. Table~\ref{table:ecuador_f} summarises our findings, where compare with SOTA method CLP on Ecuador datasets. We employ two variants of the TR with mBERT and XLM-RoBERTa architectures. Additionally, we compare with vanilla mBERT baseline. Our formulation is able to outperform both CLP and mBERT on average by $0.84$ (GNoMX), $0.75$ (GNoMM) and $0.87$ (GNoMX), $0.78$ (GNoMM) respectively.

In addition to simple crosslingual scenario, cross-disaster crosslingual settings may also arise in the disaster domain. Situations when cross-disaster crosslingual setting becomes crucial is when a classifier is trained using past disaster data in some language and applied in a different disaster with different language. 
We experiment with one such scenario in Table~\ref{table:t1_t4_f} (top two rows) for ChileEQT1 and SoSItalyT4 datasets where the classifier is trained on one disaster (e.g. earthquake) in certain language (e.g. Spanish) but evaluated on another disaster (e.g. flood + earthquake) data in another language (e.g. Italian). We compare our approach with a vanilla mBERT and able to boost the performance significantly by $11.34$ (GNoMX), $8.37$ (GNoMM) on average. We are able to achieve this performance boost by adding minimal computational complexity to TR as the additional components i.e. GF and IE contributes to only $\approx3.5\mathtt{M}$ extra parameters. For comparison, BERT-Base-Multilingual-Cased (mBERT)
alone has $\approx178\mathtt{M}$
parameters. Between XLM-RoBERTa and mBERT, XLM-RoBERTa based realization of TR performs better in crosslingual setting whereas mBERT outperforms XLM-RoBERTa in monolingual settings.
%
%
\subsection{Multilingual Classification} \label{subsec:multi_classification}
Multilingual classification setting refers to the scenario when both train and test set contains data from a mixture of multiple languages. This setting is practical in disaster scenarios as user generated social network data may be available in multiple languages.
We summarize our result for multilingual classification in Table~\ref{table:mixup_f} for the MixUp dataset. We only use the train set and do not use any extra data for construction of the word graph in this setting. However, we observe that explicit modelling of the inter-language relation (by constructing the interlanguage word graph with initial embedding similarity scores as edges) help improve performance by $1.15$ (GNoMX) and $1.29$ (GNoMM).
Unfortunately, our result can not directly be compared with~\cite{raychowdhury2020cross} as they use a larger set of data which we could not collect as it was not available (Ref.~\ref{subsec:nonenglish_datasets}).
\begin{table}[htbp]
\centering
\begin{tabular}{@{}llrrrr@{}}
\toprule
\textbf{Source}  & \textbf{Target} & \textbf{BERT} & \textbf{GNoMX} & \textbf{GNoMM} \\ \midrule
MixUp & MixUp & 70.16 & \underline{71.31} & \textbf{71.45} \\ \bottomrule
\end{tabular}
\caption{Weighted F$_{1}$ scores for MixUp (multilingual) dataset.}
\label{table:mixup_f}
\end{table}
\subsection{Limited Supervision} \label{subsec:limited_supervision}
Due to lack of labeled data in disaster domain is a common phenomenon, an effective classification system should work when very limited amount of labeled data is available. We want to verify if GNoM is capable to capture appropriate context from the unlabeled corpus so that it perform considerably well under limited supervision. To verify this, we design an experiment to limit the availability of training data to $50\%$, $25\%$ and $10\%$ of the original size, similar to~\cite{glen}.

Tables~\ref{table:limited_en_f} and~\ref{table:limited_ne_f} summarises our findings under limited supervision over the English and Non-English datasets.
We utilized the unlabeled data available with QFL, NEQ, FIRE16 and SMERP17 datasets to construct the word graph (Ref.~\ref{subsec:english_datasets}). For SoSItalyT4, ChileEQT1 and Ecuador datasets, we use the target domain data as the unlabeled data during crosslingual experiments. Note that target domain class information is not used in any of our experiments. 
For monolingual experiment, we do not use any unlabeled data.

GNoM outperforms both baseline BERT and SOTA method GLEN for English datasets by a large margin, Table~\ref{table:limited_en_f}. A vanilla BERT model overfits the small amount of training data, however, our formulation enables the model to capture larger context and overcome the overfitting problem. GLEN relies on word pair-wise contextual attention using a GAT~\cite{gat} to capture class separability, whereas our formulation uses self-attention across all the words. Additionally, our IE (cross attention) component aids in filtering noisy priors out. These additions result in average absolute gain of $2.67\%$, $3.73\%$ and $5.85\%$ with $50\%$, $25\%$ and $10\%$ of training data respectively.

For Non-English datasets, we compare with mBERT baseline only, as GLEN does not have multilingual capability. We experiment with GNoMM, additionally, we also report results on  GNoMX. We presents our experimental results over Non-English datasets in Table~\ref{table:limited_ne_f}. Our framework GNoMM consistently outperforms vanilla mBERT baseline across all training data proportions with average absolute performance gain of $3.11\%$, $3.08\%$ and $4.62\%$ with $50\%$, $25\%$ and $10\%$ of training data respectively.

\begin{table*}[htbp]
\centering
\begin{tabular}{@{}llrrrrrrrrr@{}}
\toprule
&  & \multicolumn{3}{c|}{\textbf{$50\%$}} & \multicolumn{3}{c|}{\textbf{$25\%$}} & \multicolumn{3}{c}{\textbf{$10\%$}} \\ \cmidrule(l){3-5} \cmidrule(l){6-8} \cmidrule(l){9-11}
\textbf{Source} & \textbf{Target}& \textbf{BERT} & \textbf{GLEN} & \textbf{GNoMB} & \textbf{BERT} & \textbf{GLEN} & \textbf{GNoMB} & \textbf{BERT} & \textbf{GLEN} & \textbf{GNoMB} \\ 
\midrule
NEQ & QFL & 80.49 & \underline{83.36} & \textbf{86.62} & 79.45 & \underline{82.52} & \textbf{86.12} & 78.03 & \underline{81.86} & \textbf{85.44} \\
QFL & NEQ & 65.78 & \underline{70.33} & \textbf{70.45} & 65.30 & \textbf{69.86} & \underline{69.70} & 64.62 & \textbf{69.40} & \underline{69.34} \\
NEQ & NEQ & 74.07 & \underline{75.82} & \textbf{77.77} & 71.77 & \underline{75.01} & \textbf{77.48} & 71.36 & \underline{74.19} & \textbf{75.71} \\
QFL & QFL & 94.73 & \underline{96.23} & \textbf{96.54} & 94.53 & \underline{96.14} & \textbf{96.51} & 94.09 & \underline{95.72} & \textbf{96.47} \\ 
\cmidrule(r){1-2} 
FIRE16  & SMERP17 & 72.64 & \underline{76.77} & \textbf{78.01} & 63.05 & \underline{74.67} & \textbf{74.92} & 27.42 & \underline{58.23} & \textbf{66.83} \\
SMERP17 & FIRE16  & 37.73 & \underline{47.33} & \textbf{56.82} & 28.21 & \underline{43.24} & \textbf{55.28} & 16.71 & \underline{33.81} & \textbf{47.70} \\
FIRE16  & FIRE16  & 70.44 & \underline{78.68} & \textbf{81.19} & 58.99 & \underline{72.25} & \textbf{78.43} & 42.73 & \underline{57.08} & \textbf{65.73} \\
SMERP17 & SMERP17 & 91.24 & \underline{93.49} & \textbf{95.68} & 84.03 & \underline{88.02} & \textbf{92.08} & 71.29 & \underline{77.14} & \textbf{86.38} \\
\midrule
\multicolumn{2}{c}{Average Gain (\%)} &  &  & 2.67 &  &  & 3.73 &  &  & 5.85 \\
\bottomrule
\end{tabular}
\caption{Weighted F$_{1}$ scores for NEQ, QFL, FIRE16, SMERP17, datasets under $50\%$, $25\%$ and $10\%$ of train set. We compare with baseline BERT and SOTA method GLEN. GNoM is able to outperform both.}
\label{table:limited_en_f}
\end{table*}
\begin{table*}[htbp]
\centering
\begin{tabular}{@{}llrrrrrrrrr@{}}
\toprule
&  & \multicolumn{3}{c|}{\textbf{$50\%$}} & \multicolumn{3}{c|}{\textbf{$25\%$}} & \multicolumn{3}{c}{\textbf{$10\%$}} \\ \cmidrule(l){3-5} \cmidrule(l){6-8} \cmidrule(l){9-11}
\textbf{Source} & \textbf{Target} & \textbf{mBERT} & \textbf{GNoMX} & \textbf{GNoMM} & \textbf{mBERT} & \textbf{GNoMX} & \textbf{GNoMM} & \textbf{mBERT} & \textbf{GNoMX} & \textbf{GNoMM} \\
\midrule
%
EcuadorE & EcuadorS & 77.29 & \textbf{81.47} & \underline{81.22} & 76.72 & \textbf{81.42} & \underline{81.17} & 75.40 & \textbf{81.37} & \underline{81.10} \\
EcuadorS & EcuadorE & 88.86 & \textbf{91.41} & \underline{91.30} & 87.98 & \textbf{91.22} & \underline{91.18} & 87.42 & \underline{90.92} & \textbf{91.29} \\
EcuadorE & EcuadorE & 93.84 & \underline{94.23} & \textbf{94.24} & 93.37 & \textbf{94.10} & \underline{94.07} & 92.66 & \underline{93.97} & \textbf{94.01} \\
EcuadorS & EcuadorS & 83.12 & \underline{84.36} & \textbf{84.56} & 82.24 & \underline{83.90} & \textbf{84.12} & 81.47 & \underline{83.28} & \textbf{83.57} \\ 
\cmidrule(r){1-2} 
ChileEQT1 & SoSItalyT4 & 42.50 & \textbf{51.61} & \underline{48.32} & 41.17 & \textbf{48.22} & \underline{47.20} & 22.01 & \textbf{36.80} & \underline{36.70} \\
SoSItalyT4 & ChileEQT1 & 53.38 & \textbf{66.26} & \underline{62.52} & 50.12 & \textbf{57.39} & \underline{57.32} & 46.05 & \underline{48.77} & \textbf{50.37} \\
ChileEQT1 & ChileEQT1 & 83.81 & \underline{85.49} & \textbf{85.59} & 82.38 & \textbf{83.77} & \underline{83.16} & 78.27 & \underline{81.03} & \textbf{81.78} \\ 
SoSItalyT4 & SoSItalyT4 & 84.26 & \textbf{85.13} & \underline{85.03} & 82.91 & \underline{83.89} & \textbf{84.12} & 80.47 & \underline{81.67} & \textbf{82.71} \\
\cmidrule(r){1-2} 
MixUp & MixUp & 69.49 & \underline{71.12} & \textbf{71.15} & 68.04 & \underline{69.04} & \textbf{70.32} & 65.34 & \underline{68.81} & \textbf{69.13} \\
\midrule
\multicolumn{2}{c}{Average Gain (\%)} &  &  & 3.11 &  &  & 3.08 &  &  & 4.62 \\
\bottomrule
\end{tabular}
\caption{ Weighted F$_{1}$ scores for ChileEQT1, SoSItalyT4, Ecuador and MixUp datasets under $50\%$, $25\%$ and $10\%$ of train set. We compare with multilingual BERT baseline. We use two realizations for our TR component using XLM-RoBERTa (GNoMX) and multilingual BERT (GNoMM).}
\label{table:limited_ne_f}
\end{table*}
\subsection{Ablation Study} \label{subsec:ablation}
\begin{figure*}[t!]
\centering
\subfloat[Before: SoSItalyT4-ChileEQT1]{\includegraphics[width=0.25\linewidth]{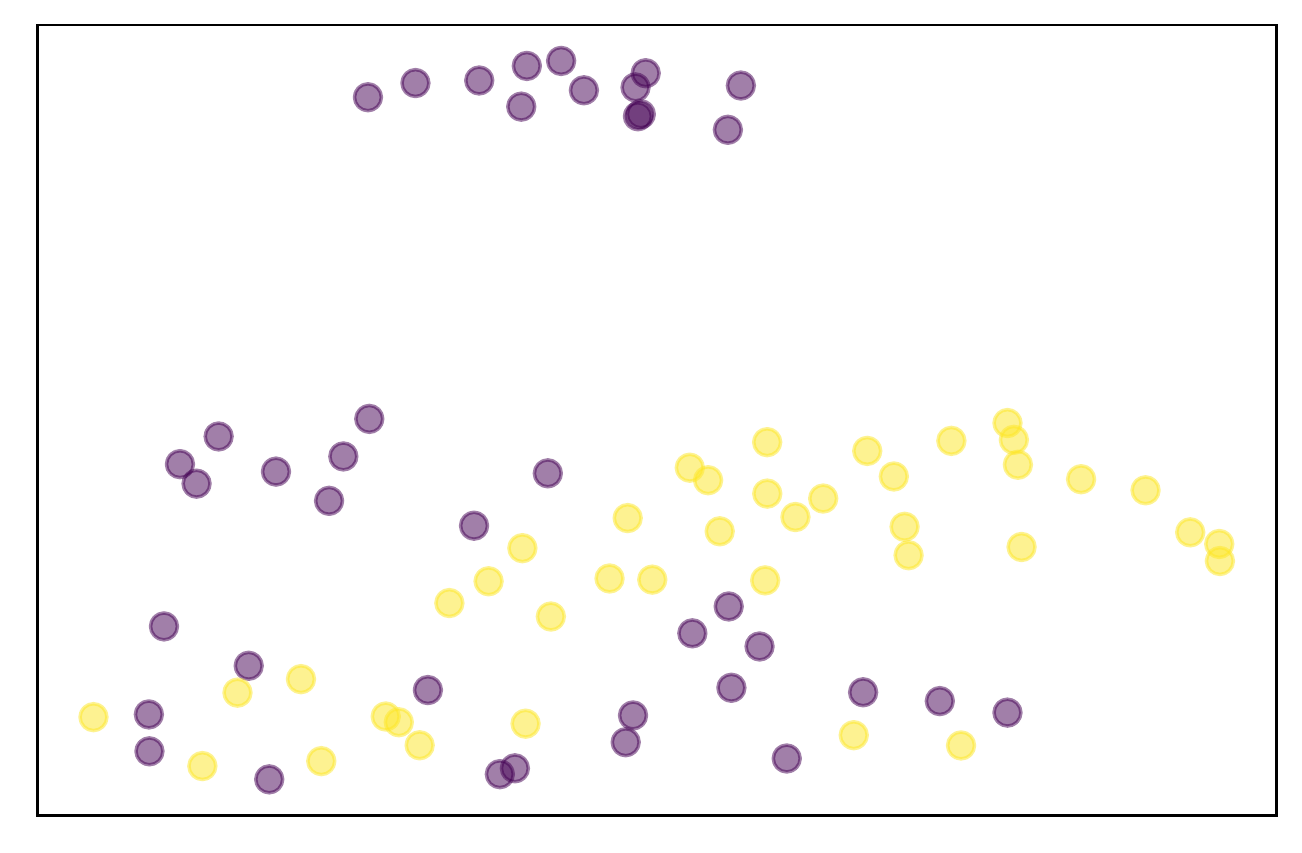}} 
\subfloat[After: SoSItalyT4-ChileEQT1]{\includegraphics[width=0.25\linewidth]{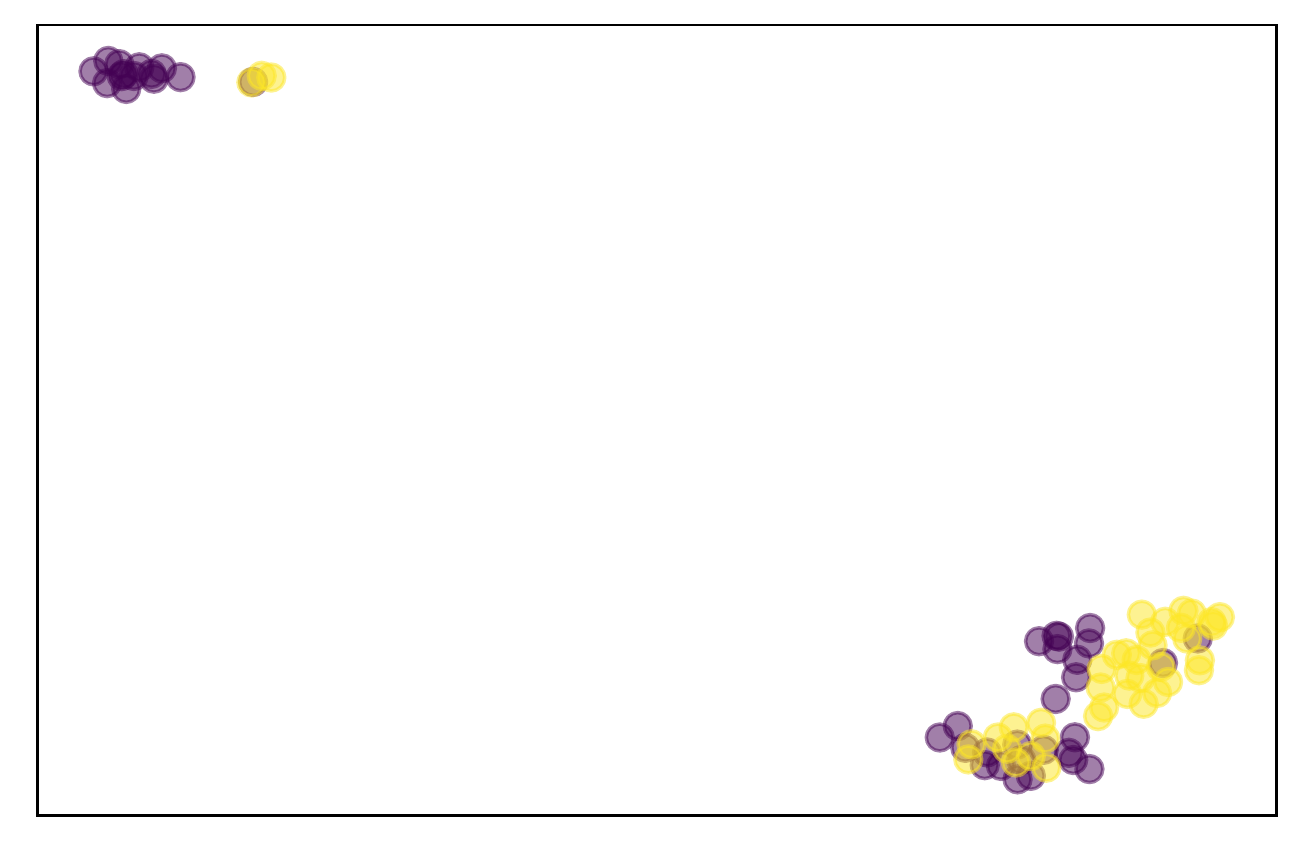}}
\subfloat[Before: EcuadorE-EcuadorS]{\includegraphics[width=0.25\linewidth]{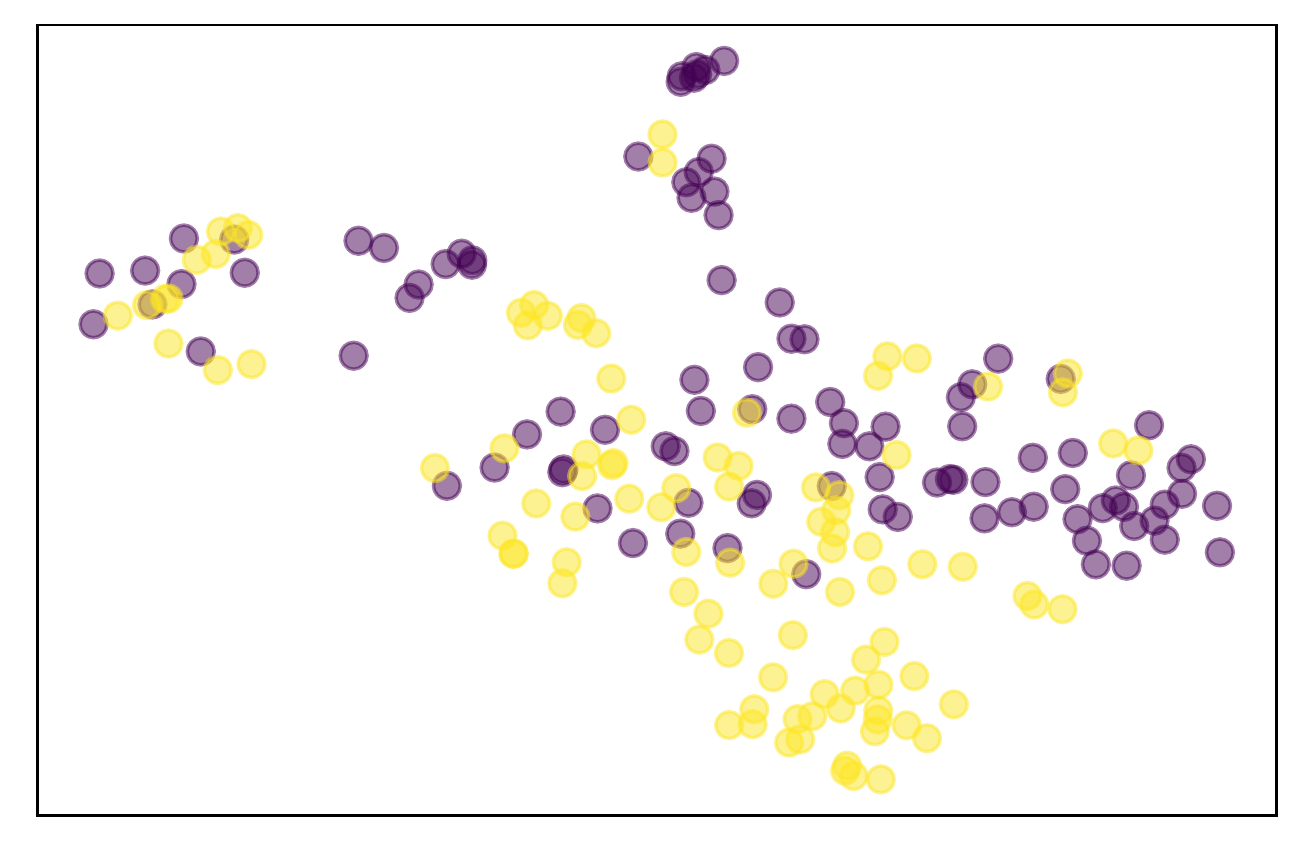}}
\subfloat[After: EcuadorE-EcuadorS]{\includegraphics[width=0.25\linewidth]{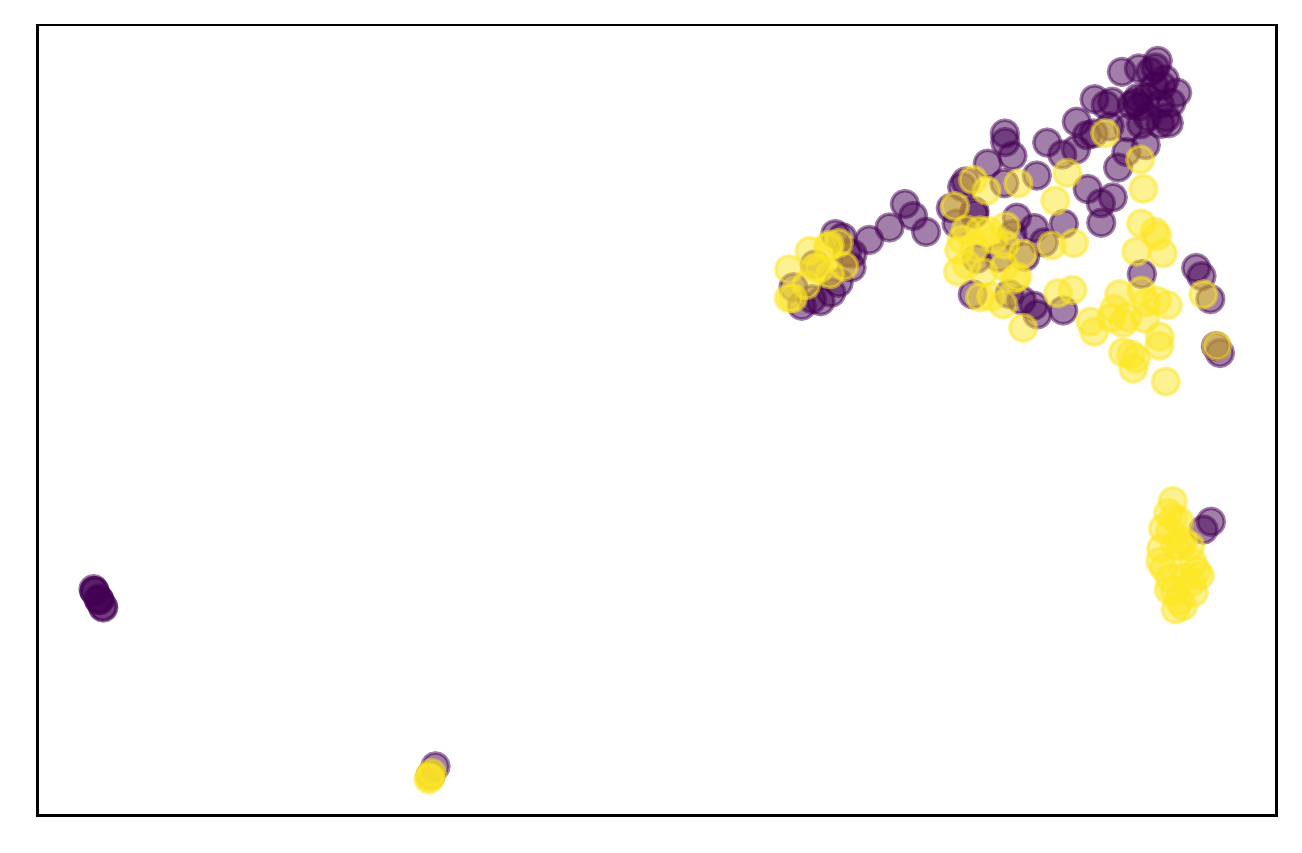}} 
\caption{UMAP projections of tokens from different languages (color-coded) before and after training. Figures (a) and (b) show for SoSItalyT4-ChileEQT1 datasets. Similarly, Fig. (c) and (d) show the plots for the EcuadorE-EcuadorS datasets. 
}
\label{fig:umap}
\end{figure*}
In the ablation study, we verify the importance of each of the components (i.e. GF, IE) within GNoM. 
For details about the model configurations, refer~\ref{subsec:baselines}.
Table~\ref{table:all_ablation} summarises the results of the ablation experiments. As evident from our experiments, incorporating GF and IE helps achieve significant performance boost over all other configurations. Our word graph based formulation is capable to capture word priors. The cross attention based IE component plays a significant role by identifying relevant prior words. 
\begin{table}[htbp]
\centering
\begin{adjustbox}{width=\linewidth}
\begin{tabular}{@{}llrrrr@{}}
\toprule
\textbf{Source} & \textbf{Target} & \textbf{TR} & \textbf{TR+GF} & \textbf{TR+GF-e+IE} & \textbf{GNoM(B|M)} \\ \midrule
NEQ & QFL & 80.72 & \underline{84.87} & - & \textbf{86.68} \\
QFL & NEQ & 67.22 & \underline{68.42} & - & \textbf{71.13} \\
\cmidrule(r){1-2}
FIRE16 & SMERP17 & 76.21 & \underline{79.37} & - &  \textbf{79.49} \\
SMERP17 & FIRE16 & 55.52 & \underline{58.90} & - & \textbf{62.57} \\
\midrule
EcuadorE & EcuadorS & 77.93 & 81.37 & \underline{81.46} & \textbf{81.54} \\
EcuadorS & EcuadorE & 90.45 & 91.25 & \underline{91.34} & \textbf{91.45} \\
\cmidrule(r){1-2}
ChileEQT1 & SoSItalyT4 & 43.17 & 46.97 & \underline{47.38} & \textbf{49.14} \\
SoSItalyT4 & ChileEQT1 & 54.46 & 56.27 & \underline{60.36} & \textbf{63.20} \\
\cmidrule(r){1-2}
MixUp & MixUp & 70.16 & 70.47 & \underline{70.68} & \textbf{71.45} \\
\midrule
\multicolumn{2}{c}{Average} & 68.08 & 70.21 & 70.24 & 72.96 \\
\bottomrule
\end{tabular}
\end{adjustbox}
\caption{Ablation with weighted F$_{1}$ scores over all the datasets. Both components (i.e. GF and IE) contribute to the improvement of performance. We experiment with TR+GF-e+IE on cross and multi lingual settings only (Ref.~\ref{subsec:training_config}). We use GNoMB for English and GNoMM for Non-English datasets.}
\label{table:all_ablation}
\end{table}

\begin{figure}[hbt]
\includegraphics[width=\linewidth] 
{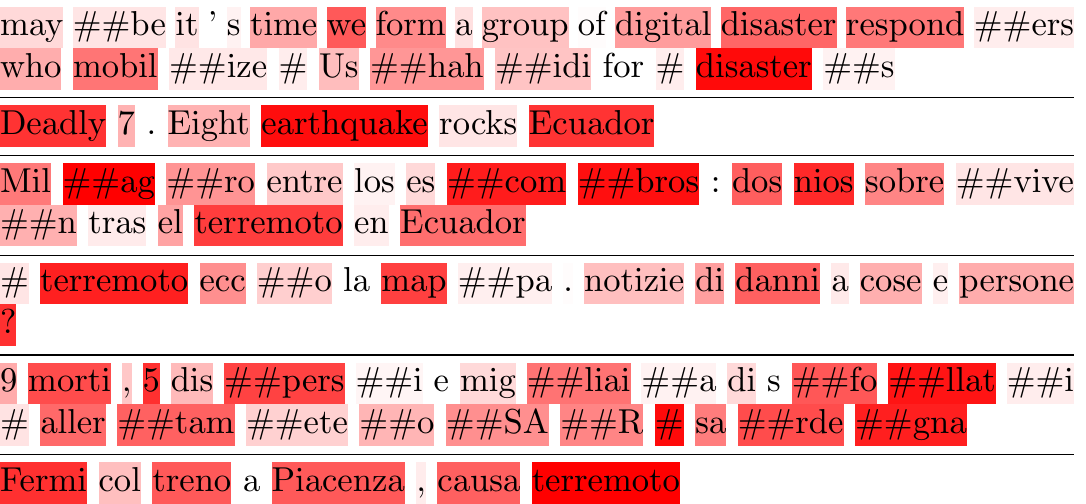}
\caption{Visualizations of cross attention scores over a few examples estimated by the IE component.} \label{fig:attention}
\end{figure}

\subsection{Qualitative Investigations} \label{subsec:quality}
We perform some manual qualitative experiments to see the effects of training on our framework.


\subsubsection{Bringing Languages Closer} \label{subsubsec:umap}
Particularly, we design an experiment to verify if words from different languages indeed come closer in the embedding space, as we assumed in our word graph formulation. We perform this experiment in crosslingual setting over SoSItalyT4 (Italian)-ChileEQT1 (Spanish) and EcuadorE (English)-EcuadorS (Spanish) datasets by randomly selecting $50$ words from each dataset (language). We obtain vectors corresponding to those words from the GF component before and after training and used  UMAP~\cite{2018arXivUMAP} projection in Figure~\ref{fig:umap}. We can clearly see words from Italian and Spanish are coming closer in the embedding space for SoSItalyT4-ChileEQT1 datasets. We also observe similar behaviour for English and Spanish language for EcuadorE-EcuadorS datasets.
\subsubsection{Cross Attention Visualization} \label{subsubsec:text_attention}
In this qualitative experiment we visualize the cross attention scores estimated by the IE component. Our cross attention formulation tries to estimate the importance of GF node vectors in respect to the example representation coming from the TR component. Figure~\ref{fig:attention} shows the cross attention scores estimated over a few examples from Ecuador and SoSItalyT4 datasets.
\subsubsection{Graph Embedding Initialization} \label{subsubsec:embs}
We mention in \S~\ref{subsection:graph_corpus_context} that we initialize the word graph using the word embedding of the TR component (i.e. transformer model). 
To verify the efficacy of these embeddings we perform a Nearest Neighbor based experiment. We selected a few words and calculated the nearest neighbors based on the initial word embedding similarity. Table~\ref{table:word_neighbors} presents the words along with $5$ nearest neighbors. This show that word embeddings in transformer based large language models contains semantic information. We utilize this semantic information as a prior in our word graph formulation.
\begin{table}[htbp]
\centering
\begin{adjustbox}{width=\linewidth}
\begin{tabular}{@{}ll@{}}
\toprule
\textbf{Word} & \textbf{Neighbors} \\ \midrule
\texttt{report} & \texttt{report, Report, report, and, port} \\
\texttt{everyone} & \texttt{everyone, Everyone, everything, anyone, people} \\
\texttt{información} & \texttt{información, informazioni, info, datos, informazio} \\
\texttt{emergencia} & \texttt{emergencia, vivienda, desastre, emergency, situación} \\
\texttt{quello} & \texttt{quello, le, vittime, das, cion} \\
\texttt{ragazzi} & \texttt{ragazzi, uomo, fer, TO, inizia} \\
\bottomrule
\end{tabular}
\end{adjustbox}
\caption{ Word and its five neighbors based on similarity of transformer word embeddings.}
\label{table:word_neighbors}
\end{table}
%
%

%
\section{Conclusion and Future Work} \label{sec:conclusion}
We proposed an multilingual disaster related text classification framework, called GNoM, which works across different languages. Explicit capturing of the corpus-level and example-level contexts enable GNoM to work under monolingual, cross-lingual and multi lingual settings. Each component of GNoM plays a crucial role to make an effective classification system at the same time being flexible with the choice of architectures. The framework is also able to work under very limited supervision significantly outperforming baselines. Our experiments over $5$ English, $3$ Non-English and $1$ multilingual datasets with binary, multi-class and multi-class multi-label settings show broader applicability of our framework in disaster related text classification.
We argue that any GNN based graph featurizer can be applied in our framework. We plan to experiment and validate this in future. We also plan to explore the possibility of applying our framework in other short-text classification domains.
%

\bibliographystyle{ACM-Reference-Format}
\bibliography{GNoM}


\begin{thebibliography}{32}


\ifx \showCODEN    \undefined \def \showCODEN     #1{\unskip}     \fi
\ifx \showDOI      \undefined \def \showDOI       #1{#1}\fi
\ifx \showISBNx    \undefined \def \showISBNx     #1{\unskip}     \fi
\ifx \showISBNxiii \undefined \def \showISBNxiii  #1{\unskip}     \fi
\ifx \showISSN     \undefined \def \showISSN      #1{\unskip}     \fi
\ifx \showLCCN     \undefined \def \showLCCN      #1{\unskip}     \fi
\ifx \shownote     \undefined \def \shownote      #1{#1}          \fi
\ifx \showarticletitle \undefined \def \showarticletitle #1{#1}   \fi
\ifx \showURL      \undefined \def \showURL       {\relax}        \fi
\providecommand\bibfield[2]{#2}
\providecommand\bibinfo[2]{#2}
\providecommand\natexlab[1]{#1}
\providecommand\showeprint[2][]{arXiv:#2}

\bibitem[Alam et~al\mbox{.}(2018)]%
        {qcri}
\bibfield{author}{\bibinfo{person}{Firoj Alam}, \bibinfo{person}{Shafiq Joty},
  {and} \bibinfo{person}{Muhammad Imran}.} \bibinfo{year}{2018}\natexlab{}.
\newblock \showarticletitle{Domain Adaptation with Adversarial Training and
  Graph Embeddings}. In \bibinfo{booktitle}{\emph{Proceedings of the 56th
  Annual Meeting of the Association for Computational Linguistics (Volume 1:
  Long Papers)}}. \bibinfo{pages}{1077--1087}.
\newblock


\bibitem[Barua et~al\mbox{.}(2021)]%
        {Barua21analysis}
\bibfield{author}{\bibinfo{person}{Aindriya Barua}, \bibinfo{person}{S. Thara},
  \bibinfo{person}{B. Premjith}, {and} \bibinfo{person}{K.~P. Soman}.}
  \bibinfo{year}{2021}\natexlab{}.
\newblock \showarticletitle{Analysis of Contextual and Non-contextual Word
  Embedding Models for Hindi NER with Web Application for Data Collection}. In
  \bibinfo{booktitle}{\emph{Advanced Computing}},
  \bibfield{editor}{\bibinfo{person}{Deepak Garg}, \bibinfo{person}{Kit Wong},
  \bibinfo{person}{Jagannathan Sarangapani}, {and}
  \bibinfo{person}{Suneet~Kumar Gupta}} (Eds.). \bibinfo{publisher}{Springer
  Singapore}, \bibinfo{address}{Singapore}, \bibinfo{pages}{183--202}.
\newblock
\showISBNx{978-981-16-0401-0}


\bibitem[Caragea et~al\mbox{.}(2016)]%
        {caragea2016identifying}
\bibfield{author}{\bibinfo{person}{Cornelia Caragea}, \bibinfo{person}{Adrian
  Silvescu}, {and} \bibinfo{person}{{Andrea H.} Tapia}.}
  \bibinfo{year}{2016}\natexlab{}.
\newblock \showarticletitle{Identifying informative messages in disaster events
  using Convolutional Neural Networks}. In \bibinfo{booktitle}{\emph{ISCRAM
  2016 Conference Proceedings - 13th International Conference on Information
  Systems for Crisis Response and Management}}
  \emph{(\bibinfo{series}{Proceedings of the International ISCRAM
  Conference})}, \bibfield{editor}{\bibinfo{person}{Pedro Antunes},
  \bibinfo{person}{{Victor Amadeo} {Banuls Silvera}}, \bibinfo{person}{Joao
  {Porto de Albuquerque}}, \bibinfo{person}{{Kathleen Ann} Moore}, {and}
  \bibinfo{person}{{Andrea H.} Tapia}} (Eds.). \bibinfo{publisher}{Information
  Systems for Crisis Response and Management, ISCRAM}.
\newblock


\bibitem[Cobo et~al\mbox{.}(2015)]%
        {cobo15chile}
\bibfield{author}{\bibinfo{person}{Alfredo Cobo}, \bibinfo{person}{Denis
  Parra}, {and} \bibinfo{person}{Jaime Nav\'{o}n}.}
  \bibinfo{year}{2015}\natexlab{}.
\newblock \showarticletitle{Identifying Relevant Messages in a Twitter-Based
  Citizen Channel for Natural Disaster Situations}. In
  \bibinfo{booktitle}{\emph{Proceedings of the 24th International Conference on
  World Wide Web}} (Florence, Italy) \emph{(\bibinfo{series}{WWW '15
  Companion})}. \bibinfo{publisher}{Association for Computing Machinery},
  \bibinfo{address}{New York, NY, USA}, \bibinfo{pages}{1189–1194}.
\newblock
\showISBNx{9781450334730}
\urldef\tempurl%
\url{https://doi.org/10.1145/2740908.2741719}
\showDOI{\tempurl}


\bibitem[Conneau et~al\mbox{.}(2019)]%
        {conneau2019unsupervised}
\bibfield{author}{\bibinfo{person}{Alexis Conneau}, \bibinfo{person}{Kartikay
  Khandelwal}, \bibinfo{person}{Naman Goyal}, \bibinfo{person}{Vishrav
  Chaudhary}, \bibinfo{person}{Guillaume Wenzek}, \bibinfo{person}{Francisco
  Guzm{\'a}n}, \bibinfo{person}{Edouard Grave}, \bibinfo{person}{Myle Ott},
  \bibinfo{person}{Luke Zettlemoyer}, {and} \bibinfo{person}{Veselin
  Stoyanov}.} \bibinfo{year}{2019}\natexlab{}.
\newblock \showarticletitle{Unsupervised Cross-lingual Representation Learning
  at Scale}.
\newblock \bibinfo{journal}{\emph{arXiv preprint arXiv:1911.02116}}
  (\bibinfo{year}{2019}).
\newblock


\bibitem[Conneau et~al\mbox{.}(2017)]%
        {conneau2017word}
\bibfield{author}{\bibinfo{person}{Alexis Conneau}, \bibinfo{person}{Guillaume
  Lample}, \bibinfo{person}{Marc'Aurelio Ranzato}, \bibinfo{person}{Ludovic
  Denoyer}, {and} \bibinfo{person}{Herv{\'e} J{\'e}gou}.}
  \bibinfo{year}{2017}\natexlab{}.
\newblock \showarticletitle{Word Translation Without Parallel Data}.
\newblock \bibinfo{journal}{\emph{arXiv preprint arXiv:1710.04087}}
  (\bibinfo{year}{2017}).
\newblock


\bibitem[Cresci et~al\mbox{.}(2015)]%
        {cresci15italy}
\bibfield{author}{\bibinfo{person}{Stefano Cresci}, \bibinfo{person}{Maurizio
  Tesconi}, \bibinfo{person}{Andrea Cimino}, {and} \bibinfo{person}{Felice
  Dell'Orletta}.} \bibinfo{year}{2015}\natexlab{}.
\newblock \showarticletitle{A Linguistically-Driven Approach to Cross-Event
  Damage Assessment of Natural Disasters from Social Media Messages}. In
  \bibinfo{booktitle}{\emph{Proceedings of the 24th International Conference on
  World Wide Web}} (Florence, Italy) \emph{(\bibinfo{series}{WWW '15
  Companion})}. \bibinfo{publisher}{Association for Computing Machinery},
  \bibinfo{address}{New York, NY, USA}, \bibinfo{pages}{1195–1200}.
\newblock
\showISBNx{9781450334730}
\urldef\tempurl%
\url{https://doi.org/10.1145/2740908.2741722}
\showDOI{\tempurl}


\bibitem[Devlin et~al\mbox{.}(2019)]%
        {bert}
\bibfield{author}{\bibinfo{person}{Jacob Devlin}, \bibinfo{person}{Ming-Wei
  Chang}, \bibinfo{person}{Kenton Lee}, {and} \bibinfo{person}{Kristina
  Toutanova}.} \bibinfo{year}{2019}\natexlab{}.
\newblock \showarticletitle{{BERT}: Pre-training of Deep Bidirectional
  Transformers for Language Understanding}. In
  \bibinfo{booktitle}{\emph{Proceedings of the 2019 Conference of the North
  {A}merican Chapter of the Association for Computational Linguistics: Human
  Language Technologies, Volume 1 (Long and Short Papers)}}.
  \bibinfo{publisher}{Association for Computational Linguistics},
  \bibinfo{address}{Minneapolis, Minnesota}, \bibinfo{pages}{4171--4186}.
\newblock
\urldef\tempurl%
\url{https://doi.org/10.18653/v1/N19-1423}
\showDOI{\tempurl}


\bibitem[Ghosh and Desarkar(2020)]%
        {ghosh2020semi}
\bibfield{author}{\bibinfo{person}{Samujjwal Ghosh} {and}
  \bibinfo{person}{Maunendra~Sankar Desarkar}.}
  \bibinfo{year}{2020}\natexlab{}.
\newblock \showarticletitle{Semi-Supervised Granular Classification Framework
  for Resource Constrained Short-texts: Towards Retrieving Situational
  Information During Disaster Events}. In \bibinfo{booktitle}{\emph{12th ACM
  Conference on Web Science}}. \bibinfo{pages}{29--38}.
\newblock


\bibitem[Ghosh and Ghosh(2016)]%
        {fire16}
\bibfield{author}{\bibinfo{person}{Saptarshi Ghosh} {and}
  \bibinfo{person}{Kripabandhu Ghosh}.} \bibinfo{year}{2016}\natexlab{}.
\newblock \showarticletitle{Overview of the {FIRE} 2016 Microblog track:
  Information Extraction from Microblogs Posted during Disasters}. In
  \bibinfo{booktitle}{\emph{Working notes of {FIRE} 2016 - Forum for
  Information Retrieval Evaluation, Kolkata, India, December 7-10, 2016}}
  \emph{(\bibinfo{series}{{CEUR} Workshop Proceedings},
  Vol.~\bibinfo{volume}{1737})}, \bibfield{editor}{\bibinfo{person}{Prasenjit
  Majumder}, \bibinfo{person}{Mandar Mitra}, \bibinfo{person}{Parth Mehta},
  \bibinfo{person}{Jainisha Sankhavara}, {and} \bibinfo{person}{Kripabandhu
  Ghosh}} (Eds.). \bibinfo{publisher}{CEUR-WS.org}, \bibinfo{pages}{56--61}.
\newblock
\urldef\tempurl%
\url{http://ceur-ws.org/Vol-1737/T2-1.pdf}
\showURL{%
\tempurl}


\bibitem[Ghosh et~al\mbox{.}(2017)]%
        {smerp17}
\bibfield{author}{\bibinfo{person}{Saptarshi Ghosh},
  \bibinfo{person}{Kripabandhu Ghosh}, \bibinfo{person}{Tanmoy Chakraborty},
  \bibinfo{person}{Debasis Ganguly}, \bibinfo{person}{Gareth Jones}, {and}
  \bibinfo{person}{Marie-Francine Moens}.} \bibinfo{year}{2017}\natexlab{}.
\newblock \showarticletitle{First International Workshop on Exploitation of
  Social Media for Emergency Relief and Preparedness (SMERP)}.
\newblock \bibinfo{journal}{\emph{ADVANCES IN INFORMATION RETRIEVAL, ECIR
  2017}}  \bibinfo{volume}{10193} (\bibinfo{year}{2017}),
  \bibinfo{pages}{779--783}.
\newblock


\bibitem[Ghosh et~al\mbox{.}(2021)]%
        {glen}
\bibfield{author}{\bibinfo{person}{Samujjwal Ghosh}, \bibinfo{person}{Subhadeep
  Maji}, {and} \bibinfo{person}{Maunendra~Sankar Desarkar}.}
  \bibinfo{year}{2021}\natexlab{}.
\newblock \bibinfo{title}{Unsupervised Domain Adaptation with Global and Local
  Graph Neural Networks in Limited Labeled Data Scenario: Application to
  Disaster Management}.
\newblock
\newblock
\showeprint[arxiv]{2104.01436}~[cs.CL]


\bibitem[Khare et~al\mbox{.}(2018)]%
        {khare2018cross}
\bibfield{author}{\bibinfo{person}{Prashant Khare},
  \bibinfo{person}{Gr{\'e}goire Burel}, \bibinfo{person}{Diana Maynard}, {and}
  \bibinfo{person}{Harith Alani}.} \bibinfo{year}{2018}\natexlab{}.
\newblock \showarticletitle{Cross-Lingual Classification of Crisis Data}. In
  \bibinfo{booktitle}{\emph{The Semantic Web -- ISWC 2018}},
  \bibfield{editor}{\bibinfo{person}{Denny Vrande{\v{c}}i{\'{c}}},
  \bibinfo{person}{Kalina Bontcheva}, \bibinfo{person}{Mari~Carmen
  Su{\'a}rez-Figueroa}, \bibinfo{person}{Valentina Presutti},
  \bibinfo{person}{Irene Celino}, \bibinfo{person}{Marta Sabou},
  \bibinfo{person}{Lucie-Aim{\'e}e Kaffee}, {and} \bibinfo{person}{Elena
  Simperl}} (Eds.). \bibinfo{publisher}{Springer International Publishing},
  \bibinfo{address}{Cham}, \bibinfo{pages}{617--633}.
\newblock
\showISBNx{978-3-030-00671-6}


\bibitem[Krishnan et~al\mbox{.}(2020)]%
        {Krishnan2020AttentionRA}
\bibfield{author}{\bibinfo{person}{Jitin Krishnan}, \bibinfo{person}{Hemant
  Purohit}, {and} \bibinfo{person}{Huzefa Rangwala}.}
  \bibinfo{year}{2020}\natexlab{}.
\newblock \showarticletitle{Attention Realignment and Pseudo-Labelling for
  Interpretable Cross-Lingual Classification of Crisis Tweets}. In
  \bibinfo{booktitle}{\emph{KiML@KDD}}.
\newblock


\bibitem[Li and Caragea(2020)]%
        {sigir'20}
\bibfield{author}{\bibinfo{person}{Xukun Li} {and} \bibinfo{person}{Doina
  Caragea}.} \bibinfo{year}{2020}\natexlab{}.
\newblock \showarticletitle{Domain Adaptation with Reconstruction for Disaster
  Tweet Classification}. In \bibinfo{booktitle}{\emph{Proceedings of the 43rd
  International ACM SIGIR Conference on Research and Development in Information
  Retrieval}}. \bibinfo{pages}{1561--1564}.
\newblock


\bibitem[Lorini et~al\mbox{.}(2019)]%
        {lorini2019integrating}
\bibfield{author}{\bibinfo{person}{V. Lorini}, \bibinfo{person}{C. Castillo},
  \bibinfo{person}{F. Dottori}, \bibinfo{person}{M. Kalas}, \bibinfo{person}{D.
  Nappo}, {and} \bibinfo{person}{P. Salamon}.} \bibinfo{year}{2019}\natexlab{}.
\newblock \bibinfo{title}{Integrating Social Media into a Pan-European Flood
  Awareness System: A Multilingual Approach}.
\newblock
\newblock
\showeprint[arxiv]{1904.10876}~[cs.IR]


\bibitem[Mazloom et~al\mbox{.}(2018)]%
        {Mazloom2018ClassificationOT}
\bibfield{author}{\bibinfo{person}{Reza Mazloom}, \bibinfo{person}{Hongming
  Li}, \bibinfo{person}{Doina Caragea}, \bibinfo{person}{Muhammad Imran}, {and}
  \bibinfo{person}{Cornelia Caragea}.} \bibinfo{year}{2018}\natexlab{}.
\newblock \showarticletitle{Classification of Twitter Disaster Data Using a
  Hybrid Feature-Instance Adaptation Approach}. In
  \bibinfo{booktitle}{\emph{ISCRAM}}.
\newblock


\bibitem[{McInnes} et~al\mbox{.}(2018)]%
        {2018arXivUMAP}
\bibfield{author}{\bibinfo{person}{L. {McInnes}}, \bibinfo{person}{J. {Healy}},
  {and} \bibinfo{person}{J. {Melville}}.} \bibinfo{year}{2018}\natexlab{}.
\newblock \showarticletitle{{UMAP: Uniform Manifold Approximation and
  Projection for Dimension Reduction}}.
\newblock \bibinfo{journal}{\emph{ArXiv e-prints}} (\bibinfo{date}{Feb.}
  \bibinfo{year}{2018}).
\newblock
\showeprint[arxiv]{1802.03426}~[stat.ML]


\bibitem[Mikolov et~al\mbox{.}(2013)]%
        {word2vec}
\bibfield{author}{\bibinfo{person}{Tomas Mikolov}, \bibinfo{person}{Ilya
  Sutskever}, \bibinfo{person}{Kai Chen}, \bibinfo{person}{Greg~S Corrado},
  {and} \bibinfo{person}{Jeff Dean}.} \bibinfo{year}{2013}\natexlab{}.
\newblock \showarticletitle{Distributed Representations of Words and Phrases
  and their Compositionality}. In \bibinfo{booktitle}{\emph{Advances in Neural
  Information Processing Systems}}, \bibfield{editor}{\bibinfo{person}{C.~J.~C.
  Burges}, \bibinfo{person}{L.~Bottou}, \bibinfo{person}{M.~Welling},
  \bibinfo{person}{Z.~Ghahramani}, {and} \bibinfo{person}{K.~Q. Weinberger}}
  (Eds.), Vol.~\bibinfo{volume}{26}. \bibinfo{publisher}{Curran Associates,
  Inc.}, \bibinfo{pages}{3111--3119}.
\newblock
\urldef\tempurl%
\url{https://proceedings.neurips.cc/paper/2013/file/9aa42b31882ec039965f3c4923ce901b-Paper.pdf}
\showURL{%
\tempurl}


\bibitem[Musaev and Pu(2017)]%
        {Musaev2017TowardsMA}
\bibfield{author}{\bibinfo{person}{Aibek Musaev} {and} \bibinfo{person}{Calton
  Pu}.} \bibinfo{year}{2017}\natexlab{}.
\newblock \showarticletitle{Towards Multilingual Automated Classification
  Systems}.
\newblock \bibinfo{journal}{\emph{2017 IEEE 37th International Conference on
  Distributed Computing Systems (ICDCS)}} (\bibinfo{year}{2017}),
  \bibinfo{pages}{2333--2337}.
\newblock


\bibitem[Neppalli et~al\mbox{.}(2018)]%
        {Neppalli2018DeepNN}
\bibfield{author}{\bibinfo{person}{Venkata~Kishore Neppalli},
  \bibinfo{person}{Cornelia Caragea}, {and} \bibinfo{person}{Doina Caragea}.}
  \bibinfo{year}{2018}\natexlab{}.
\newblock \showarticletitle{Deep Neural Networks versus Naive Bayes Classifiers
  for Identifying Informative Tweets during Disasters}. In
  \bibinfo{booktitle}{\emph{ISCRAM}}.
\newblock


\bibitem[Nguyen et~al\mbox{.}(2017)]%
        {Nguyen2017RobustCO}
\bibfield{author}{\bibinfo{person}{Tien~Dat Nguyen}, \bibinfo{person}{Kamla
  Al-Mannai}, \bibinfo{person}{Shafiq~R. Joty}, \bibinfo{person}{Hassan
  Sajjad}, \bibinfo{person}{Muhammad Imran}, {and} \bibinfo{person}{Prasenjit
  Mitra}.} \bibinfo{year}{2017}\natexlab{}.
\newblock \showarticletitle{Robust Classification of Crisis-Related Data on
  Social Networks Using Convolutional Neural Networks}. In
  \bibinfo{booktitle}{\emph{ICWSM}}.
\newblock


\bibitem[Oono and Suzuki(2020)]%
        {oono2020graph}
\bibfield{author}{\bibinfo{person}{Kenta Oono} {and} \bibinfo{person}{Taiji
  Suzuki}.} \bibinfo{year}{2020}\natexlab{}.
\newblock \showarticletitle{Graph Neural Networks Exponentially Lose Expressive
  Power for Node Classification}. In \bibinfo{booktitle}{\emph{International
  Conference on Learning Representations}}.
\newblock
\urldef\tempurl%
\url{https://openreview.net/forum?id=S1ldO2EFPr}
\showURL{%
\tempurl}


\bibitem[Piscitelli et~al\mbox{.}(2021)]%
        {Piscitelli2021multilingual}
\bibfield{author}{\bibinfo{person}{Sara Piscitelli}, \bibinfo{person}{Edoardo
  Arnaudo}, {and} \bibinfo{person}{Claudio Rossi}.}
  \bibinfo{year}{2021}\natexlab{}.
\newblock \showarticletitle{Multilingual Text Classification from Twitter
  during Emergencies}. In \bibinfo{booktitle}{\emph{2021 IEEE International
  Conference on Consumer Electronics (ICCE)}}. \bibinfo{pages}{1--6}.
\newblock
\urldef\tempurl%
\url{https://doi.org/10.1109/ICCE50685.2021.9427581}
\showDOI{\tempurl}


\bibitem[Ray~Chowdhury et~al\mbox{.}(2020)]%
        {raychowdhury2020cross}
\bibfield{author}{\bibinfo{person}{Jishnu Ray~Chowdhury},
  \bibinfo{person}{Cornelia Caragea}, {and} \bibinfo{person}{Doina Caragea}.}
  \bibinfo{year}{2020}\natexlab{}.
\newblock \showarticletitle{Cross-Lingual Disaster-related Multi-label Tweet
  Classification with Manifold Mixup}. In \bibinfo{booktitle}{\emph{Proceedings
  of the 58th Annual Meeting of the Association for Computational Linguistics:
  Student Research Workshop}}. \bibinfo{publisher}{Association for
  Computational Linguistics}, \bibinfo{address}{Online},
  \bibinfo{pages}{292--298}.
\newblock
\urldef\tempurl%
\url{https://doi.org/10.18653/v1/2020.acl-srw.39}
\showDOI{\tempurl}


\bibitem[Torres(2019)]%
        {torres2019cross}
\bibfield{author}{\bibinfo{person}{Johnny Torres, Carmen~Vaca}.}
  \bibinfo{year}{2019}\natexlab{}.
\newblock \showarticletitle{Cross-Lingual Perspectives about Crisis-Related
  Conversations on Twitter}. In \bibinfo{booktitle}{\emph{Companion Proceedings
  of The 2019 World Wide Web Conference}} (San Francisco, USA)
  \emph{(\bibinfo{series}{WWW '19})}. \bibinfo{publisher}{Association for
  Computing Machinery}, \bibinfo{address}{New York, NY, USA},
  \bibinfo{pages}{255–261}.
\newblock
\showISBNx{9781450366755}
\urldef\tempurl%
\url{https://doi.org/10.1145/3308560.3316799}
\showDOI{\tempurl}


\bibitem[Veli{\v{c}}kovi{\'c} et~al\mbox{.}(2018)]%
        {gat}
\bibfield{author}{\bibinfo{person}{Petar Veli{\v{c}}kovi{\'c}},
  \bibinfo{person}{Guillem Cucurull}, \bibinfo{person}{Arantxa Casanova},
  \bibinfo{person}{Adriana Romero}, \bibinfo{person}{Pietro Li{\`o}}, {and}
  \bibinfo{person}{Yoshua Bengio}.} \bibinfo{year}{2018}\natexlab{}.
\newblock \showarticletitle{Graph Attention Networks}. In
  \bibinfo{booktitle}{\emph{Proceedings of the 6th International Conference on
  Learning Representations}} \emph{(\bibinfo{series}{ICLR '18})}.
\newblock


\bibitem[Verma et~al\mbox{.}(2019)]%
        {Verma19manifoldmixup}
\bibfield{author}{\bibinfo{person}{Vikas Verma}, \bibinfo{person}{Alex Lamb},
  \bibinfo{person}{Christopher Beckham}, \bibinfo{person}{Amir Najafi},
  \bibinfo{person}{Ioannis Mitliagkas}, \bibinfo{person}{David Lopez-Paz},
  {and} \bibinfo{person}{Yoshua Bengio}.} \bibinfo{year}{2019}\natexlab{}.
\newblock \showarticletitle{Manifold Mixup: Better Representations by
  Interpolating Hidden States}. In \bibinfo{booktitle}{\emph{Proceedings of the
  36th International Conference on Machine Learning}}
  \emph{(\bibinfo{series}{Proceedings of Machine Learning Research},
  Vol.~\bibinfo{volume}{97})}, \bibfield{editor}{\bibinfo{person}{Kamalika
  Chaudhuri} {and} \bibinfo{person}{Ruslan Salakhutdinov}} (Eds.).
  \bibinfo{publisher}{PMLR}, \bibinfo{address}{Long Beach, California, USA},
  \bibinfo{pages}{6438--6447}.
\newblock
\urldef\tempurl%
\url{http://proceedings.mlr.press/v97/verma19a.html}
\showURL{%
\tempurl}


\bibitem[Wang et~al\mbox{.}(2021)]%
        {wang2021transformer}
\bibfield{author}{\bibinfo{person}{Congcong Wang}, \bibinfo{person}{Paul
  Nulty}, {and} \bibinfo{person}{David Lillis}.}
  \bibinfo{year}{2021}\natexlab{}.
\newblock \showarticletitle{Transformer-based Multi-task Learning for Disaster
  Tweet Categorisation}.
\newblock \bibinfo{journal}{\emph{arXiv preprint arXiv:2110.08010}}
  (\bibinfo{year}{2021}).
\newblock


\bibitem[Wu et~al\mbox{.}(2020)]%
        {Wu_Lian_Xu_Wu_Chen_2020}
\bibfield{author}{\bibinfo{person}{Yongji Wu}, \bibinfo{person}{Defu Lian},
  \bibinfo{person}{Yiheng Xu}, \bibinfo{person}{Le Wu}, {and}
  \bibinfo{person}{Enhong Chen}.} \bibinfo{year}{2020}\natexlab{}.
\newblock \showarticletitle{Graph Convolutional Networks with Markov Random
  Field Reasoning for Social Spammer Detection}.
\newblock \bibinfo{journal}{\emph{Proceedings of the AAAI Conference on
  Artificial Intelligence}} \bibinfo{volume}{34}, \bibinfo{number}{01}
  (\bibinfo{date}{Apr.} \bibinfo{year}{2020}), \bibinfo{pages}{1054--1061}.
\newblock
\urldef\tempurl%
\url{https://doi.org/10.1609/aaai.v34i01.5455}
\showDOI{\tempurl}


\bibitem[Yao et~al\mbox{.}(2019)]%
        {yao2019graph}
\bibfield{author}{\bibinfo{person}{Liang Yao}, \bibinfo{person}{Chengsheng
  Mao}, {and} \bibinfo{person}{Yuan Luo}.} \bibinfo{year}{2019}\natexlab{}.
\newblock \showarticletitle{Graph convolutional networks for text
  classification}. In \bibinfo{booktitle}{\emph{Proceedings of the AAAI
  conference on artificial intelligence}}, Vol.~\bibinfo{volume}{33}.
  \bibinfo{pages}{7370--7377}.
\newblock


\bibitem[Zahera et~al\mbox{.}(2021)]%
        {iaid}
\bibfield{author}{\bibinfo{person}{Hamada~M. Zahera}, \bibinfo{person}{Rricha
  Jalota}, \bibinfo{person}{Mohamed~Ahmed Sherif}, {and}
  \bibinfo{person}{Axel-Cyrille~Ngonga Ngomo}.}
  \bibinfo{year}{2021}\natexlab{}.
\newblock \showarticletitle{I-AID: Identifying Actionable Information From
  Disaster-Related Tweets}.
\newblock \bibinfo{journal}{\emph{IEEE Access}}  \bibinfo{volume}{9}
  (\bibinfo{year}{2021}), \bibinfo{pages}{118861--118870}.
\newblock
\urldef\tempurl%
\url{https://doi.org/10.1109/ACCESS.2021.3107812}
\showDOI{\tempurl}


\end{thebibliography}










\end{document}